\documentclass[10pt,twocolumn,letterpaper]{article}

\usepackage{cvpr}      
\usepackage{times}
\usepackage{epsfig}
\usepackage{graphicx}
\usepackage{amsmath}
\usepackage{amssymb}
\usepackage{color}
\usepackage{pifont}
\usepackage{booktabs}
\usepackage{multirow}

\usepackage[hang,flushmargin]{footmisc}

\usepackage[accsupp]{axessibility} 

\usepackage[pagebackref=true,breaklinks=true,letterpaper=true,colorlinks,bookmarks=True]{hyperref}

\def\cvprPaperID{92} 
\def\confName{CVPR}
\def\confYear{2023}

\begin{document}
	
	\title{NTIRE 2023 Challenge on Light Field Image Super-Resolution:\\
	Dataset, Methods and Results}
	\author{Yingqian Wang$^{*}$, Longguang Wang$^{*}$, Zhengyu Liang$^{*}$, Jungang Yang$^{* \dag}$, Radu Timofte$^{*}$, Yulan Guo$^{*}$,\\
	Kai Jin, Zeqiang Wei, Angulia Yang, Sha Guo, Mingzhi Gao, Xiuzhuang Zhou, Vinh Van Duong,\\
	Thuc Nguyen Huu, Jonghoon Yim, Byeungwoo Jeon, Yutong Liu, Zhen Cheng, Zeyu Xiao,\\
	Ruikang Xu, Zhiwei Xiong, Gaosheng Liu, Manchang Jin, Huanjing Yue, Jingyu Yang, Chen Gao,\\
	Shuo Zhang, Song Chang, Youfang Lin, Wentao Chao, Xuechun Wang, Guanghui Wang,\\
	Fuqing Duan, Wang Xia, Yan Wang, Peiqi Xia, Shunzhou Wang, Yao Lu, Ruixuan Cong, Hao Sheng,\\
	Da Yang, Rongshan Chen, Sizhe Wang, Zhenglong Cui, Yilei Chen, Yongjie Lu,
	Dongjun Cai,\\
    Ping An, Ahmed Salem, Hatem Ibrahem, Bilel Yagoub, Hyun-Soo Kang, Zekai Zeng, Heng Wu
	}
	
	\maketitle

	\begin{abstract}
	In this report, we summarize the first NTIRE challenge on light field (LF) image super-resolution (SR), which aims at super-resolving LF images under the standard bicubic degradation with a magnification factor of 4. This challenge develops a new LF dataset called NTIRE-2023 for validation and test, and provides a toolbox called BasicLFSR to facilitate model development. Compared with single image SR, the major challenge of LF image SR lies in how to exploit complementary angular information from plenty of views with varying disparities. In total, 148  participants have registered the challenge, and 11 teams have successfully submitted results with PSNR scores higher than the baseline method LF-InterNet \cite{LF-InterNet}. These newly developed methods have set new state-of-the-art in LF image SR, e.g., the winning method achieves around 1 dB PSNR improvement over the existing state-of-the-art method DistgSSR \cite{DistgLF}. We report the solutions proposed by the participants, and summarize their common trends and useful tricks. We hope this challenge can stimulate future research and inspire new ideas in LF image SR. 
	\end{abstract}
	\vspace{-0.2cm}
	

	\footnotetext{
	\noindent $^{*}$Yingqian Wang, Longguang Wang, Zhengyu Liang, Jungang Yang, Radu Timofte and Yulan Guo are the NTIRE 2023 challenge organizers, while the other authors participated in this challenge.	\\
	\noindent $^\dag$Corresponding author: Jungang Yang\\
    ~~Section \ref{appendix} provides the authors and affiliations of each team.\\
	~~NTIRE 2023 webpage: \url{https://cvlai.net/ntire/2023/}\\
	~~Challenge webpage: \url{https://codalab.lisn.upsaclay.fr/competitions/9201}\\
	~~Leaderboard: \url{https://codalab.lisn.upsaclay.fr/competitions/9201\#results}\\
	~~Github: \url{https://github.com/The-Learning-And-Vision-Atelier-LAVA/LF-Image-SR/tree/NTIRE2023}\\
	~~BasicLFSR toolbox: \url{https://github.com/ZhengyuLiang24/BasicLFSR}
 }
 
 \section{Introduction}
 
 Light field (LF) cameras can capture both intensity and directions of light rays, and record 3D geometry in a convenient and efficient manner. By encoding 3D scene cues into 4D LF images (i.e., 2D for spatial dimension and 2D for angular dimension), LF cameras enable many attractive applications such as post-capture refocusing \cite{vaish2004using,wang2018selective}, depth sensing \cite{SPO,CAE,EPINET,LFAttNet,AttMLFNet,OACC-Net,SubFocal,LFNAT2023-LFDE}, virtual reality \cite{overbeck2018system,yu2017light} and view rendering \cite{wu2021revisiting,sitzmann2021light,wang2022r2l,attal2022learning}.
 
 In many applications, high-resolution (HR) LF images are highly demanded to achieve higher perceptual quality and benefit downstream applications. However, HR LF images are generally obtained at an expensive cost due to the spatial-angular trade-off issue in LF imaging \cite{zhu2019revisiting}. Consequently, it is highly necessary to reconstruct HR LF images from their low-resolution (LR) counterparts, i.e., to achieve LF image super-resolution (SR).

 In recent years, remarkable progress has been achieved in image SR with deep learning techniques. However, most approaches focus on super-resolving single images \cite{SRCNN14,DASR,SMSR,MSRN,RCAN,Lu2022Transformer}, stereo images \cite{PASSRnet,iPASSR,Dai2021Feedback,Chu2022NAFSSR,guo2023pftssr} or videos \cite{EDVR,D3Dnet,BasicVSR,SOFVSR20}, and cannot be directly extended to the task of LF image SR. For LF images, how to effectively incorporate both spatial and angular information is important but challenging.
 
 To develop and benchmark LF image SR methods, we host the first LF image SR challenge on the NTIRE 2023 workshop. This challenge employs the widely used and publicly available LF datasets \cite{EPFL,HCInew,HCIold,INRIA,STFgantry} as training set, and proposes a new LF dataset called NTIRE-2023 for both validation (model development) and test (final ranking). The popular bicubic degradation is used to generate LR LF images, and the objective of this challenge is to make the super-resolved LF images as faithful as the groundtruth HR ones. Besides, this challenge provides an open-source and easy-to-use toolbox named BasicLFSR to facilitate participants to quickly get access to LF image SR and develop their own models. In summary, this challenge aims at establishing a new benchmark for LF image SR, and aspires to highlight specific challenges and research problems. We hope that this challenge can inspire the community to explore the cross area of low-level vision and 3D vision, and stimulate future research in LF image processing.
 

 This challenge is one of the NTIRE 2023 Workshop~\footnote{https://cvlai.net/ntire/2023/} series of challenges on: night photography rendering~\cite{shutova2023ntire_night}, HR depth from images of specular and transparent surfaces~\cite{zama2023ntire_depth}, image denoising~\cite{li2023ntire_dn50}, video colorization~\cite{kang2023ntire_vc}, shadow removal~\cite{vasluianu2023ntire_isr}, quality assessment of video enhancement~\cite{liu2023ntire}, stereo super-resolution~\cite{wang2023ntire_ssr}, light field image super-resolution~\cite{wang2023ntire_lfsr}, image super-resolution ($\times4$)~\cite{zhang2023ntire}, 360° omnidirectional image and video super-resolution~\cite{cao2023ntire}, lens-to-lens bokeh effect transformation~\cite{conde2023ntire_bokeh}, real-time 4K super-resolution~\cite{conde2023ntire_rtsr}, HR nonhomogenous dehazing~\cite{ancuti2023ntire}, efficient super-resolution~\cite{li2023ntire_esr}.

 \section{Related Work}\label{SecRelatedWork}
 In this section, we briefly review several major works in LF image SR. We divide existing LF image SR methods into traditional non-learning methods, CNN-based methods and Transformer-based methods. Note that, we only focus on the plain-lens based methods, and do not discuss those hybird-lens based LF image SR methods \cite{chang2022flexible,jin2020light,zheng2017combining,wang2016light,chen2022deep}.

 \subsection{Traditional Methods}

 Light field image SR is a long-standing problem and has been investigated for decades. Bishop et al. \cite{bishop2011light} proposed a Bayesian deconvolution approach to super-resolve LF images based on the estimated disparities. Wanner et al. \cite{wanner2013variational} first estimated disparity maps using structure tensor, and then developed a variational framework for LF image SR. Farrugia et al. \cite{Farrugia2017} constructed a patch-volume dictionary of HR-LR LF image pairs, and proposed a multivariate ridge regression method to learn the linear mapping from LR patch volumes to their HR counterparts. In \cite{alain2017light}, Alain et al. considered the ill-posed LF image SR problem as an optimization problem based on the sparsity prior. Rossi et al. \cite{rossi2017graph} combined the inter-view information using graph regularization, and formulated LF image SR as a quadratic problem which can be solved efficiently with standard convex optimization. 

 \subsection{CNN-based Methods}

 In the past decade, convolutional neural networks (CNNs) have been extensively studied and achieved remarkable performance in LF image SR. Yoon et al. \cite{LFCNN15} proposed the first CNN-based LF image SR method (i.e., LFCNN). In their method, input LF images were grouped into pairs or quads, and fed to a three layer CNN to integrate complementary information from adjacent views. As the pioneering work, LFCNN \cite{LFCNN15} shows great potential of CNNs in LF image SR. Afterwards, many deeper CNNs with various angular information incorporation mechanisms were developed to achieve improved SR performance. 

 Wang et al. \cite{LFNet} proposed a bidirectional recurrent CNN (i.e., LFNet) to incorporate angular information from the sub-aperture images (SAIs) along the horizontal or vertical angular direction. Zhang et al. \cite{resLF} stacked SAIs along four different angular directions, and developed a four-branch residual network to implicitly learn the epipolar geometry from stacked SAIs for LF image SR. In their subsequent work, Zhang et al. \cite{MEG-Net} improved the SR performance by performing 3D convolutions on SAI stacks of different angular directions. Cheng et al. \cite{cheng2019light} developed a framework to exploit both internal and external similarities for LF image SR. Meng et al. \cite{HDDRNet} applied 4D convolutions to simultaneously incorporate spatial and angular information from 4D LF data, and developed a high-dimensional dense residual network (HDDRNet) for LF image SR. Jin et al. \cite{ATO} proposed an all-to-one method for LF image SR, and performed structural consistency regularization to preserve the parallax structure. Wang et al. \cite{LF-DFnet} applied deformable convolution to LF spatial SR, and designed a collect-and-distribute scheme to incorporate the complementary information among different views. Mo et al. \cite{DDAN} proposed a dense dual-attention network (DDAN) for LF image SR, in which a view attention module and a channel attention module were designed to adaptively capture discriminative information from different views and channels, respectively. 
 
 Instead of directly processing 4D LF data or image stacks, some methods disentangled 4D LFs into different subspace for LF image SR. Yeung et al. \cite{LFSSR} alternately reshaped LF images between SAI pattern and macro-pixel pattern, and designed spatial-angular separable convolutions for LF image SR. In \cite{LF-InterNet}, Wang et al. proposed spatial and angular feature extractors to extract corresponding information from macro-pixel images (MacPIs), and developed an LF-InterNet to repetitively interact the spatial and angular information for LF image SR. In their subsequent work, Wang et al. \cite{DistgLF} further generalized the interaction mechanism into LF disentangling mechanism, and developed three CNNs (i.e., DistgSSR, DistgASR and DistgDisp) for spatial SR, angular SR and disparity estimation, respectively.  Following \cite{LF-InterNet}, Liu et al. \cite{LF-IINet} proposed an intra-inter view interaction network (LF-IINet) with two parallel branches to extract global inter-view information and model the correlations among all intra-view features, respectively. These two branches are mutually interacted to fuse angular and spatial information for LF image SR.

 Besides the aforementioned works that design advanced network structures to pursuit superior SR accuracy, several works also studied some special yet important issue in LF image SR. Cheng et al. \cite{ZSLFSR} addressed the domain gap issue by proposing a “zero-shot” learning framework, in which the network learns to achieve spatial SR without using external training data except the given input LR LF. Wang et al. \cite{LF-DAnet} addressed the degradation formulation issue in LF image SR, and proposed a method to handle LF image SR with multiple degradation. Xiao et al. \cite{CutMIB} proposed a data augmentation approach tailored for LF image SR, which can be applied to existing LF image SR networks to further improve their SR performance

 \subsection{Transformer-based Methods}
 
 Transformer networks, which were originally developed for natural language processing \cite{Vaswani2017Attention}, have recently gained much attention in computer vision community. Recently, Transformers have been successfully applied to many low-level vision tasks such as image restoration \cite{SwinIR,ipt, Lu2022Transformer} and video SR \cite{RVRT,VRT,VSRT}, and achieved superior performance than CNN-based methods.

 In the past two years, researchers have explored Transformers for LF image SR. Wang et al. \cite{DPT} proposed a detail-preserving Transformer (DPT) for LF image SR, in which SAIs of each vertical and horizontal views are considered as a sequence, and the long-range geometric dependency is learned via a spatial-angular locally enhanced self-attention layer. Liang et al. \cite{LFT} proposed a simple yet effective Transformer network (i.e., LFT) for LF image SR. In their method, an angular Transformer is designed to incorporate complementary information among different views, and a spatial Transformer is developed to capture both local and long-range dependencies within each SAI. Guo et al. \cite{guo2022light} develop a raw LF data generation pipeline to utilize the rich information from the raw LF data to enhance their spatial resolution. They introduced a volume Transformer to aggregate information of all views into center view, and designed a cross-view Transformer to align the center view feature to all views for non-local information utilization. Wang et al. \cite{wang2022multi} proposed a Multi-granularity Aggregation Transformer (MAT) for LF image SR, in which the LF feature representation was learned via three designed granularity aggregation units. More recently, Liang et al. \cite{EPIT} investigated the non-local spatial-angular correlations in LF image SR, and developed a Transformer-based network called EPIT to achieve state-of-the-art SR performance. The proposed EPIT achieves a global receptive field along the epipolar line, and is robust to disparity variations.

\section{NTIRE 2023 Challenge}
In this section, we introduce the NTIRE 2023 LF image SR Challenge. We first introduce the datasets used in this challenge, and then briefly describe the BasicLFSR toolbox. Afterwards, we review the two phases of this challenge, and finally summarize the common trends in the submitted solutions.

\begin{figure*}[t]
     \centering
     \includegraphics[width=1\linewidth]{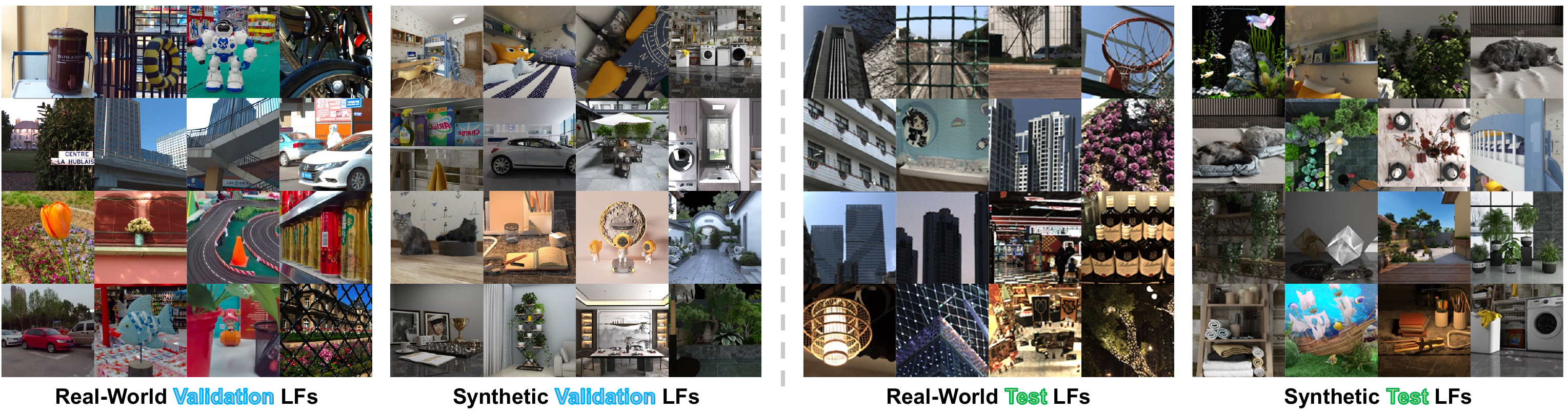}
     \caption{An illustration of the center-view images in the developed NTIRE-2023 LF dataset. Both validation and test sets contain 16 real-world and 16 synthetic LFs, respectively.}
     \label{fig:datasets}
 \end{figure*}

\subsection{Dataset}
\noindent \textbf{Training Set.}
This challenge follows the existing LF image SR works \cite{LF-DFnet,DistgLF, DPT, LFT, LF-IINet, EPIT}, and uses the EPFL \cite{EPFL}, HCInew \cite{HCInew}, HCIold \cite{HCIold}, INRIA \cite{INRIA} and STFgantry \cite{STFgantry} datasets for training. All the 144 LFs in the training set have an angular resolution of $9\times9$. The participants are required to use these LF images as HR groundtruth to train their models. External training data or models pretrained on other datasets are not allowed in this challenge.

\noindent \textbf{Validation Set.} 
 In this challenge, we develop a new LF dataset (namely, NTIRE-2023) for both validation and test, as shown in Fig.~\ref{fig:datasets}. The validation set contains 16 synthetic scenes rendered by the 3DS MAX software\textcolor[RGB]{230, 0, 0}{$^{\text{1}}$} and 16 real-world images captured by Lytro Illum cameras. For synthetic scenes, all virtual cameras in the camera array have identical internal parameters and are co-planar with the parallel optical axes.  All scenes in the validation set have an angular resolution of $5\times 5$. The spatial resolutions of synthetic LFs and real-world LFs are $500\times500$ and $624\times432$, respectively. All the LF images in the validation set are bicubicly downsampled by a factor of 4, and only the LR versions are released to the participants. Challenge participants are required to apply their developed models to the LR LF images, and submit the super-resolved LF images to the CodaLab server for validation. 

\footnotetext[1]{\textcolor[RGB]{230, 0, 0}{$^{\text{1}}$}https://www.autodesk.eu/products/3ds-max/overview}

\noindent \textbf{Test Set.}
To rank the submitted models, a new test set consisting of 16 synthetic LFs (rendered in the same way as in the validation set) and 16 real-world LFs (captured by Lytro Illum cameras) are provided, as shown in Fig.~\ref{fig:datasets}. Same as the validation set, only 4$\times$ downsampled LR LF images with an angular resolution of 5$\times$5 are released to the participants. 

 \subsection{The BasicLFSR Toolbox}
 This challenge provides a PyTorch-based, open-source, and easy-to-use toolbox named BasicLFSR to facilitate participants to quickly get access to LF image SR and develop their own models. The BasicLFSR toolbox has the following three characteristics: (1) It provides a complete pipeline to develop novel LF image SR methods. (2) It integrates a number of LF image SR methods, and retrains them on unified LF datasets. The codes and checkpoints of each model are publicly available. (3) It provides a fair and comprehensive benchmark for LF image SR. The quantitative results of each method are listed, and their super-resolved LF images are available for download.

\subsection{Challenge Phases}
\noindent {\textbf{Development Phase.}} 
The participants can download the LR validation set and apply their developed models to the LR LF images to generate their SR versions. A validation leaderboard is available online, and the participants can compare their scores with the ones achieved by the baseline models (provided by the challenge organizers) or models developed by other participants. 
	
\noindent {\textbf{Test phase.}} 
The participants are required to apply their models to the released LR test set, and submit their super-resolved LF images to the test server. The test server is available online during this phase, and will be closed after the test deadline. The participants are asked to submit the SR results, codes and a fact sheet of their methods before the given deadlines. After this challenge, the final rank is released to the participants, and the test server will be re-open to facilitate the development of novel LF image SR methods in the future. 

\noindent \textbf{Evaluation Metrics.}
Peak signal-to-noise ratio (PSNR) and structural similarity (SSIM) are used as metrics for performance evaluation. The implementation details of PSNR and SSIM can be found in the BasicLFSR toolbox. The submitted results are ranked by the average PSNR values on the test set (both real-world and synthetic scenes). 

\begin{table*}[t]
		\caption{NTIRE 2023 LF Image SR Challenge results, final rankings, and the main characteristics of the solutions. Note that, the average PSNR value achieved on the test set is used for final ranking. The best results are in \textcolor[RGB]{230, 0, 0}{{\bf red}}, the second best results are in \textcolor[RGB]{0, 47, 163}{\bf blue}, and the third best results are in \textcolor[RGB]{0, 150, 0}{\bf green}.}
		\label{tab1}
		\centering
		\renewcommand\arraystretch{1.2}
		\scriptsize
		\setlength{\tabcolsep}{1.0mm}{
			\begin{tabular}{clccccccrccccc}
				\hline
				\multirow{2}*{Rank} 
				& \multirow{2}*{Team}
				& \multicolumn{3}{c}{Test Set}
				& \multicolumn{3}{c}{Validation Set}
                & \multirow{2}*{\#Params}
				& \multirow{2}*{Architec*}
				& \multirow{2}*{Subspace}
				& \multirow{2}*{Ensemble}
				\tabularnewline
				\cline{3-8}
				& 
				& Average
				& Lytro
				& Synthetic
				& Average
				& Lytro
				& Synthetic
				& & & & \tabularnewline
				\hline
                1 & OpenMeow\textcolor[RGB]{255,215,0}{$^\bigstar$}      
                & \textcolor[RGB]{230, 0, 0}{{\bf 30.66}}$/$\textcolor[RGB]{0, 150, 0}{\bf .9314} & \textcolor[RGB]{0, 47, 163}{\bf 30.82}$/$\textcolor[RGB]{0, 150, 0}{\bf .9475} & \textcolor[RGB]{230, 0, 0}{{\bf 30.51}}$/$\textcolor[RGB]{0, 47, 163}{\bf .9152} 
                & \textcolor[RGB]{230, 0, 0}{{\bf 32.71}}$/$\textcolor[RGB]{230, 0, 0}{{\bf .9496}} & \textcolor[RGB]{230, 0, 0}{{\bf 33.36}}$/$\textcolor[RGB]{0, 47, 163}{\bf .9562} & \textcolor[RGB]{230, 0, 0}{{\bf 32.07}}$/$\textcolor[RGB]{230, 0, 0}{{\bf .9430}} & 20.34M 
                & Hybrid & Spa \& Ang \& EPI & Data \& Model
				\tabularnewline
    
				2 & DMLab\textcolor[RGB]{192,192,192}{$^{\bigstar}$}        
                & \textcolor[RGB]{0, 47, 163}{\bf 30.64}$/$\textcolor[RGB]{0, 47, 163}{\bf .9318} & \textcolor[RGB]{230, 0, 0}{{\bf 30.92}}$/$\textcolor[RGB]{0, 47, 163}{\bf .9489} & \textcolor[RGB]{0, 150, 0}{\bf 30.35}$/$\textcolor[RGB]{0, 150, 0}{\bf .9146} 
                & \textcolor[RGB]{0, 150, 0}{\bf 32.43}$/$\textcolor[RGB]{0, 150, 0}{\bf .9485} & \textcolor[RGB]{0, 47, 163}{\bf 33.24}$/$\textcolor[RGB]{0, 150, 0}{\bf .9559}  & 31.62$/$\textcolor[RGB]{0, 150, 0}{\bf .9410} 
                & 28.99M & CNN & Spa \& Ang \& EPI & Data
				\tabularnewline

                3 & VIDAR\textcolor[RGB]{184,115,51}{$^{\bigstar}$}        
                & \textcolor[RGB]{0, 150, 0}{\bf 30.56}$/$\textcolor[RGB]{230, 0, 0}{{\bf .9323}} & \textcolor[RGB]{0, 150, 0}{\bf 30.67}$/$\textcolor[RGB]{230, 0, 0}{{\bf .9491}} & \textcolor[RGB]{0, 47, 163}{\bf 30.45}$/$\textcolor[RGB]{230, 0, 0}{{\bf .9154}} 
                & \textcolor[RGB]{0, 47, 163}{\bf 32.54}$/$\textcolor[RGB]{0, 47, 163}{\bf .9494} & \textcolor[RGB]{0, 47, 163}{\bf 33.24}$/$\textcolor[RGB]{230, 0, 0}{{\bf .9568}} & \textcolor[RGB]{0, 47, 163}{\bf 31.85}$/$\textcolor[RGB]{0, 47, 163}{\bf .9419}
                & 10.52M & Transf & Spa \& Ang \& EPI & Data \& Model
                \tabularnewline

                4 & IIR-Lab      
                & 30.38$/$.9285 & 30.56$/$.9450 & 30.20$/$.9119 & 32.24$/$.9465 & 32.84$/$.9529 & \textcolor[RGB]{0, 150, 0}{\bf 31.64}$/$.9402
                & 2.63M & Transf & Spa \& Ang \& EPI & - 
                \tabularnewline

                5 & INSIS        
                & 30.35$/$.9287 & 30.56$/$.9458 & 30.15$/$.9117 & 32.12$/$.9455 & 32.86$/$.9526 & 31.39$/$.9383
                & 5.43M & CNN & Spa \& Ang \& EPI & Data
                \tabularnewline

                6 & BNU-AI-TRY    
                & 30.13$/$.9290 & 29.97$/$.9453 & 30.29$/$.9126 & 32.29$/$.9468 & 32.96$/$.9539 & 31.63$/$.9396 
                & 8.83M & Transf & Spa \& EPI & Data \& Model
                \tabularnewline

                7 & BIT912       
                & 30.11$/$.9293 & 30.10$/$.9465 & 30.13$/$.9120 & 32.05$/$.9449 & 32.76$/$.9528 & 31.35$/$.9371
                & 4.08M & Transf & Spa \& Ang \& EPI & -
                \tabularnewline

                8 & HawkeyeGroup  & 30.06$/$.9285 & 29.99$/$.9447 & 30.13$/$.9124 & 32.13$/$.9463 & 32.86$/$.9543 & 31.40$/$.9383
                & 3.35M & Transf & Spa \& Ang & -
                \tabularnewline

                9 & SHU-IVIPLab   & 29.90$/$.9265 & 29.78$/$.9433 & 30.01$/$.9096 & 32.01$/$.9442 & 32.69$/$.9517 & 31.32$/$.9366
                & 7.79M & CNN & Spa \& Ang \& EPI & Data
                \tabularnewline

                10 & CBNU-MIP-Lab   & 29.85$/$.9279 & 29.64$/$.9447 & 30.06$/$.9111 & 32.13$/$.9464 & 32.70$/$.9533 & 31.55$/$.9395
                & 14.82M & CNN & Spa \& EPI & -
                \tabularnewline

                11 & LFSR-gdut-team & 29.83$/$.9262 & 29.64$/$.9422 & 30.01$/$.9103 & 31.83$/$.9431 & 32.53$/$.9508 & 31.13$/$.9354
                & 7.28M & CNN & Spa \& EPI & -
                \tabularnewline


				\hline
				- &  EPIT \cite{EPIT}                & 29.87$/$.9259 & 29.72$/$.9420 & 30.03$/$.9097 & 32.04$/$.9447 & 32.54$/$.9507 & 31.53$/$.9387 
                & 1.47M & Transf & Spa \& EPI & \ding{55}
				\tabularnewline

				- &  LFT \cite{LFT}                  & 29.77$/$.9252 & 29.66$/$.9420 & 29.88$/$.9084 & 31.75$/$.9423 & 32.42$/$.9501 & 31.08$/$.9344 
                & 1.16M & Transf & Spa \& Ang & \ding{55}
				\tabularnewline

				- &  DistgSSR \cite{DistgLF}         & 29.64$/$.9244 & 29.39$/$.9403 & 29.88$/$.9084 & 31.75$/$.9424 & 32.26$/$.9490 & 31.23$/$.9357 
                & 3.58M & CNN & Spa \& Ang \& EPI & \ding{55}
				\tabularnewline

				- &  LF-InterNet \cite{LF-InterNet}  & 29.45$/$.9198 & 29.23$/$.9369 & 29.45$/$.9028 & 31.33$/$.9381 & 32.06$/$.9468 & 30.61$/$.9295 
                & 5.48M & CNN & Spa \& Ang & \ding{55}
				\tabularnewline

				- &  Bicubic                         & 25.79$/$.8378 & 25.11$/$.8404 & 26.46$/$.8352 & 27.51$/$.8714 & 27.49$/$.8719 & 27.53$/$.8710 
                & -~~~~~ & \ding{55} & Spa & \ding{55} 
				\tabularnewline

				\hline
		\end{tabular}}
		\vspace{0.2cm}
		\begin{tabular}{l}
		 \leftline{~~Note: ``Transf'' denotes that the model adopts Transformer as a basic component, ``CNN'' denotes that the model was developed based on convolutions only. }  \\
		 \leftline{~~~~~~~~~~~~~ ``Hybird'' denotes that the model contains sub-models which are developed based on CNNs and Transformers, respectively.}  \\
		\end{tabular}
	\end{table*}

\subsection{Challenge Results}
 Among the 148 registered participants, 12 teams have successfully participated the final test phase and submitted their results, codes, and factsheets. The top 11 of them produced PSNR scores higher than the baseline method LF-InterNet \cite{LF-InterNet}. Table~\ref{tab1} reports the PSNR and SSIM scores achieved by these methods on both test and validation sets, together with their major details. We briefly describe these solutions in Section \ref{sec:Methods}, and introduce the corresponding team members in Appendix \ref{appendix}.
 
 It can be observed from Table~\ref{tab1} that all these methods surpass the state-of-the-art method DistgSSR \cite{DistgLF}, and 9 of them surpass the recent top-performing method EPIT \cite{EPIT}. Note that, the winner solution proposed by the OpenMeow achieves around 1 dB improvement in PSNR over DistgSSR \cite{DistgLF} on both test and validation sets, which significantly push the state-of-the-art of LF image SR to a new height. Moreover, the accuracy of the top 2 methods are very close with a minor PSNR difference of 0.02 dB on the test set. In addition, although the second runner-up solution proposed by the VIDAR team produces slightly inferior PSNR results than the winner solution and the runner-up solution, it achieves the highest SSIM score of 0.9323 on the test set.

 \noindent \textbf{Architectures and main ideas.}
 All the proposed methods are based on deep learning techniques. Transformers are used as the basic architecture in 6 solutions, while other models are purely based on CNNs. The idea of LF disentangling \cite{DistgLF} was adopted in most solutions, and the recently developed method EPIT \cite{EPIT} was used as the backbone by the OpenMeow team (winner) and the BNU-AI-TRY team.
	
 \noindent \textbf{Subspace division.}
 Since an LF has a complex structure and its spatial and angular information
 is highly coupled with varying disparities, it is challenging for deep neural networks to exploit informative cues from such a high-dimensional tensor. Consequently, 7 teams adopted the disentangling mechanism in \cite{DistgLF} to divide the 4D LFs into four 2D subspaces including spatial subspace (i.e., SAIs), angular subspace (i.e., macro-pixels), horizontal EPI subspace, and vertical EPI subspace. Three teams performed feature extraction and incorporation in spatial and EPI subspaces, while one team learned LF image SR in spatial and angular subspaces.

 \noindent \textbf{Data Augmentation.}
 The participants commonly performed random flipping and rotation for training data augmentation. In addition, two teams randomly sampled $5\times 5$ LFs from $9\times 9$ LFs to further augment the training set. However, some advanced data augmentation approaches such as CutBlur \cite{CutBlur} and RGB channel shuffling have not been adopted in this challenge.
    
 \noindent \textbf{Ensemble Strategy.}
 Both data ensemble (a.k.a. test-time augmentation) and model ensemble were adopted in several solutions to boost the SR performance. For data ensemble~\cite{timofte2016seven}, the inputs were flipped and rotated, and the resultant SR images were aligned and averaged for enhanced prediction. Note that, the INSIS team proposed a shear ensemble approach tailored with LF image SR for performance enhancement. The OpenMeow, VIDAR, and BNU-AI-TRY teams adopted model ensemble, and averaged the results produced by multiple models for better results.

 \noindent \textbf{Conclusions.}
 By analyzing the settings, the proposed methods and their results, we can conclude that:
 \begin{itemize}
 \item The proposed solutions significantly improve the state-of-the-art in LF image SR.
 \item Transformers are increasingly popular in LF image SR, but the well-designed CNNs (e.g., the solution proposed by the DMLab team) can also achieve competitive SR performance.
 \item Most methods exploring multi-dimensional information from spatial, angular and EPI subspaces. Spatial and EPI subspaces are quite important for achieving competitive SR performance.
 \item There seems to be a considerable room of further performance improvement, because ensemble strategy and some advanced data augmentation approaches have not been widely used. 
 \end{itemize}

\section{Challenge Teams and Methods}
\label{sec:Methods}

\subsection{OpenMeow: DistgEPIT\textcolor[RGB]{255,215,0}{$^\bigstar$}}

\begin{figure*}[t]
    \centering
    \includegraphics[width=0.95\linewidth]{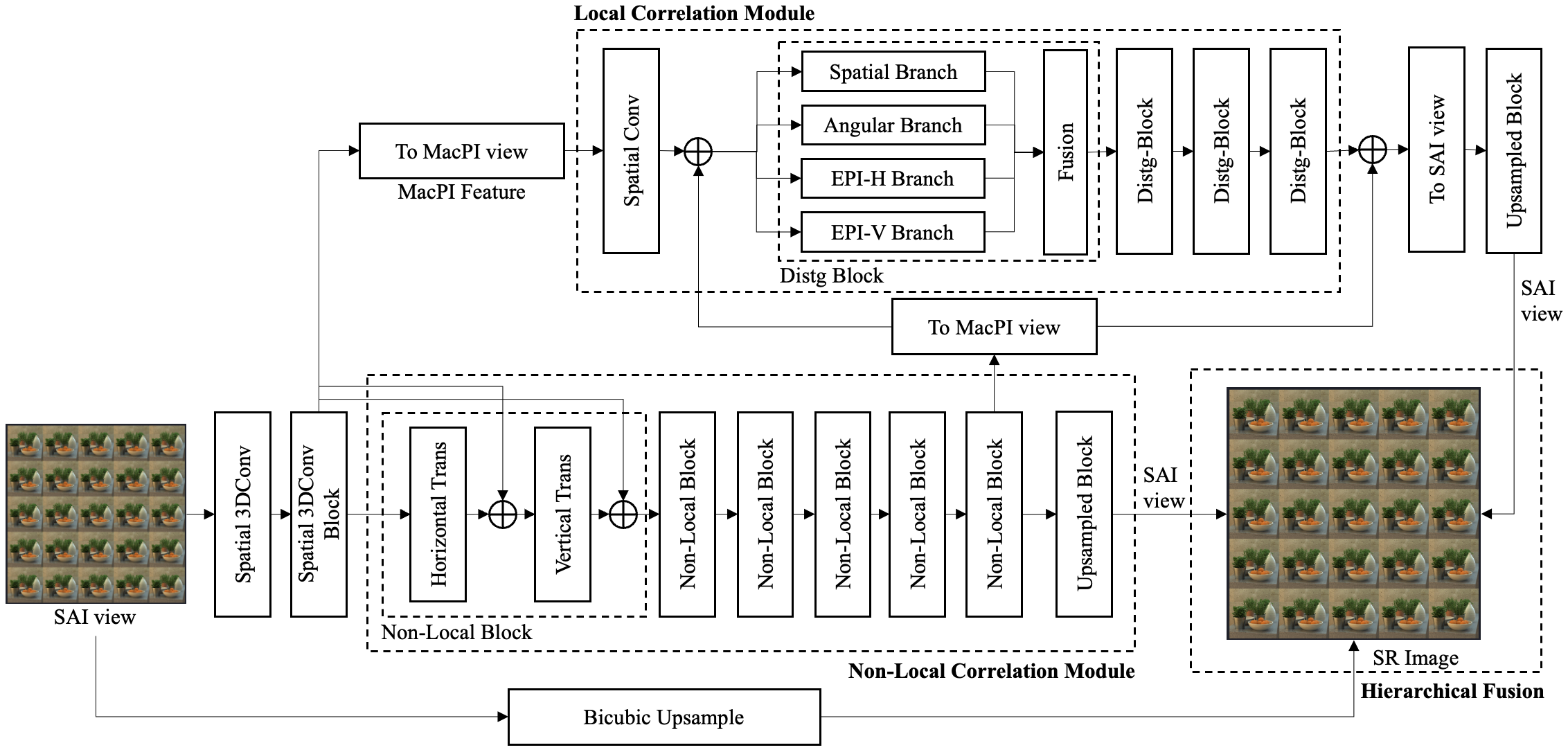}
    \caption{The OpenMeow Team: The network architecture of the proposed DistgEPIT.} 
    \label{fig:DistgEPIT}
\end{figure*}

The OpenMeow team proposed a hybrid network called DistgEPIT for LF image SR. Readers can refer to \cite{DistgEPIT} for more details of their method. The proposed DistgEPIT contains a DistgSSR-based branch \cite{DistgLF} and an EPIT-based branch \cite{EPIT}, which can learn the spatial-angular relationship from the MacPI representation while handling the large disparity issue by adopting the EPI representation. As shown in Fig.~\ref{fig:DistgEPIT}, the DistgEPIT network adopts the non-local cascading block (i.e., Basic-Transformer unit in EPIT \cite{EPIT}) to exploit information from sub-aperture images (SAIs) along the horizontal and vertical angular directions. The long-range modeling ability of the non-local cascading block benefits the learning of pixel-wise correlations from remote views. After extracting deep features via several non-local cascading blocks, the OpenMeow team uses several Distg-Blocks \cite{DistgLF} for refinement. The final SR results are generated by fusing the bicubicly upsampled image, the output of the EPIT branch, and the output of the DistgSSR branch.

Moreover, this team proposed a position-sensitive post-processing method to eliminate the margin of LF patches introduced by the commonly used zero-padding in the \textit{LF divide-and-integrate operation}\textcolor[RGB]{230, 0, 0}{$^{\text{2}}$}. Specifically, they adopted a sliding window approach to crop the chop in an overlapping manner without introducing any padding operations. 
In cases where the last row or last column is cropped, the window backtracks to make up the entire chop.

\footnote{\textcolor[RGB]{230, 0, 0}{$^{\text{2}}$}Please refer to the BasicLFSR toolbox for the implementation details.}

\noindent \textbf{Ensemble Strategy:} 
The OpenMeow team performed model ensemble by using three different configurations of DistgEPIT and two different configurations of DistgSSR. Specifically, in the first DistgEPIT model (i.e., DistgEPIT\_wider), each local correlation module has 128 channels and includes 4 Distg-Groups (each Distg-Group has 4 Distg-Blocks). The second configuration of DistgEPIT, called DistgEPIT\_deeper, has 64 channels but increases the number of non-local correlation blocks from 5 to 8. Moreover, the number of Distg-Groups in the local correlation module is increased from 4 to 8. The third configuration of DistgEPIT, called DistgEPIT\_Parallel, extracts features from both local and non-local correlation modules in parallel, and fuses MacPIs at the top level using two cascaded Distg-Groups (each Distg-Group has two Distg-Blocks). The two configurations of DistgSSR have 64 and 128 channels, respectively, and the convolution kernels in the original upsampling layer of DistgSSR are modified from $1\times1$ to $3\times3$. In total, 12 groups of model parameters were obtained from different training phases. For data augmentation, horizontal flip, vertical flip, and 90-degree rotation were used, and the final results were obtained by aligning and averaging the results of all models and data.


\subsection{DMLab: RR-HLFSR\textcolor[RGB]{192,192,192}{$^{\bigstar}$}}

\begin{figure*}[t]
    \centering
    \includegraphics[width=1\linewidth]{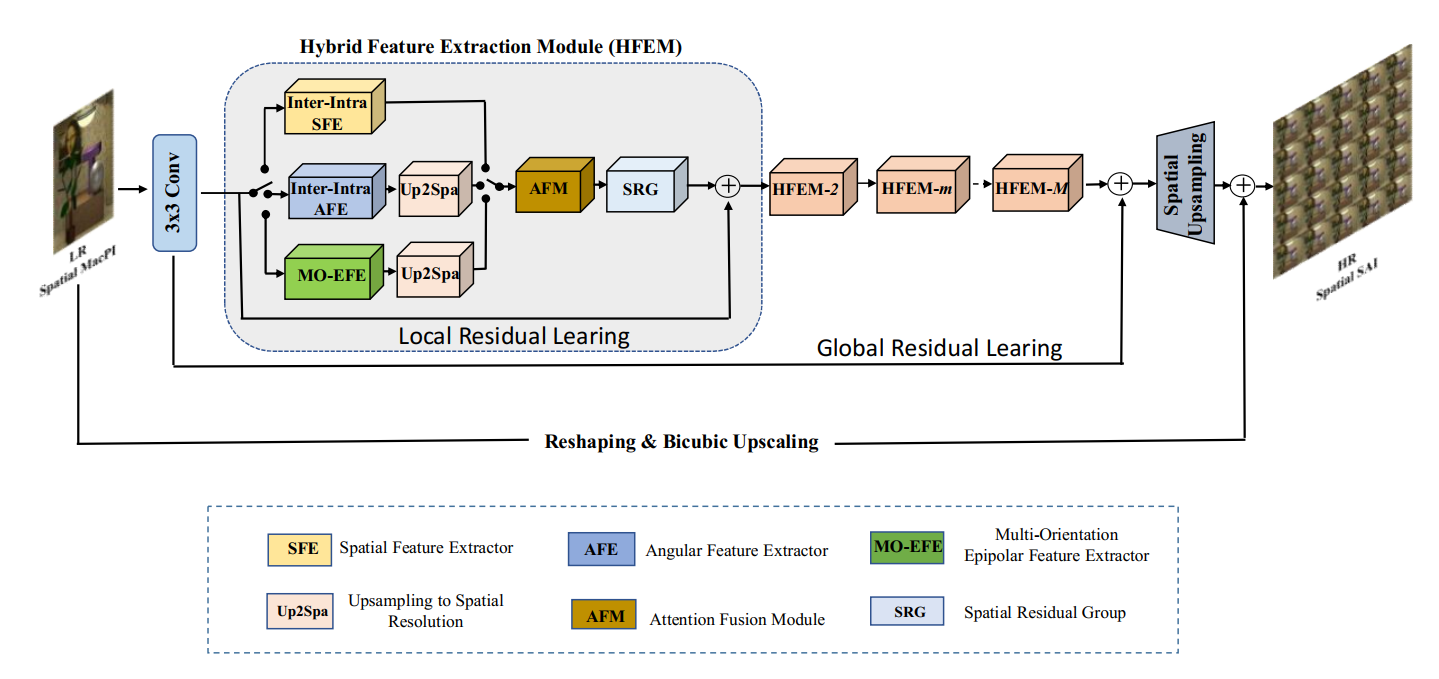}
    \caption{The DMLab Team: The network architecture of the proposed RR-HLFSR.} 
    \label{fig:RR-HLFSR}
\end{figure*}

This team presented a residual in residual learning based hybrid LF image SR network (namely, RR-HLFSR), which is an enhanced version of their recently published method HLFSR \cite{HLFSR}. The main improvement of RR-HLFSR as compared to HLFSR is that the local residual learning and global residual learning are introduced to the basic hybrid feature extraction, as shown in Fig.~\ref{fig:RR-HLFSR}. Thanks to the residual learning mechanism, the RR-HLFSR network can be developed deeper than HLFSR, and achieves considerable improvements in SR performance. 

The proposed RR-HLFSR network contains three types of 2D feature extractors that work in different sub-spaces of 4D LFs: Inter-Intra Spatial Feature Extractor (II-SFE), Inter-Intra Angular Feature Extractor (II-AFE), and Multi-Orientation Epipolar Feature Extractor (MO-EFE). Specifically, the II-SFE and II-AFE are designed to explore the correlation among pixels within each SAI and each macro-pixel, respectively. The MO-EFE is designed to handle multiple stacks of SAIs with different epipolar geometry orientations to extract abundant sub-pixel information. 

Moreover, since diverse information can be extracted from multiple sub-spaces, how to effectively fuse various features from different feature extractors is crucial in further improving the quality of recovered LF images.  
This method designed an attention fusion module (AFM) that handles fused information from different branches. By using the simple but effective modules, the SR performance is enhanced.

\subsection{VIDAR: SAVformer\textcolor[RGB]{184,115,51}{$^{\bigstar}$}}

\begin{figure*}[t]
    \centering
    \includegraphics[width=1\linewidth]{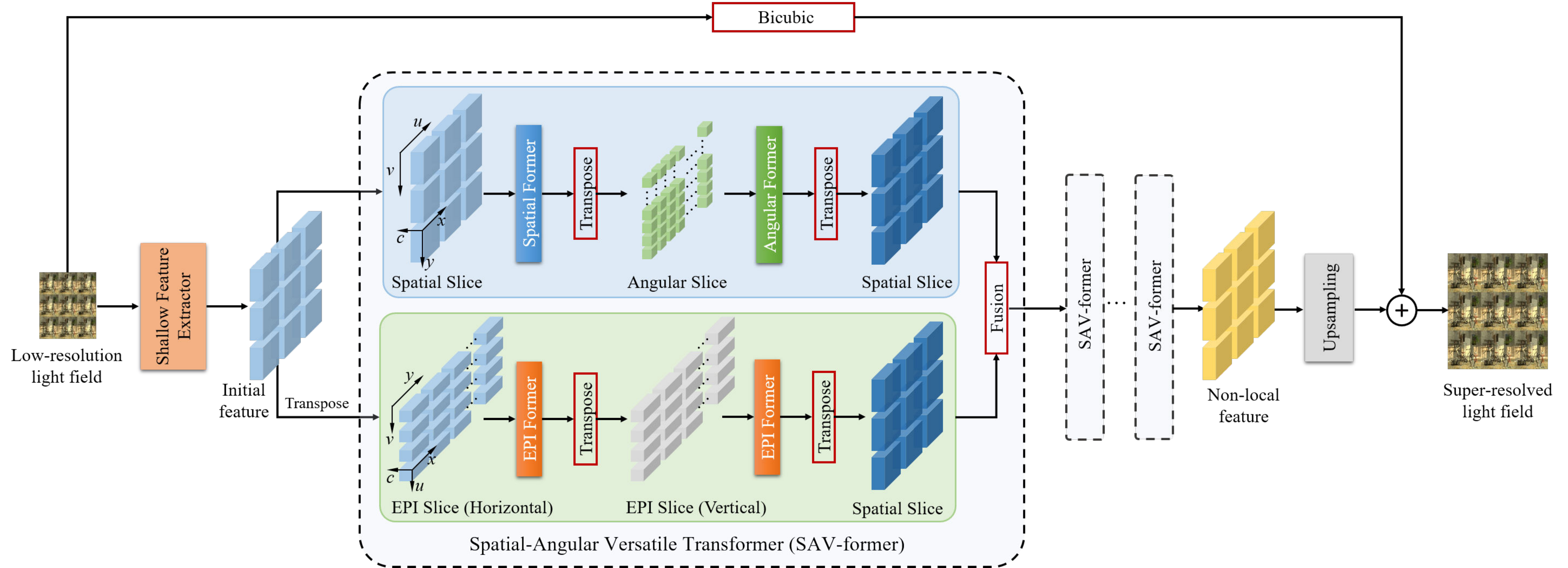}
    \caption{The VIDAR Team: The network architecture of the proposed SAVformer.} 
    \label{fig:vidar}
\end{figure*}

This method is mainly inspired by their published work LFSSR-SAV \cite{LFSSR_SAV} and the mile-stone single image restoration method Swin-Transformer \cite{SwinIR}. In LFSSR-SAV \cite{LFSSR_SAV}, the authors proposed a novel spatial-angular correlated convolution (SAC-conv) and adopted the spatial-angular separable convolution (SAS-conv) \cite{LFSSR} for efficient LF feature extraction, and verified that both SAS-conv and SAC-conv are complementary at different aspects of 4D LF feature embedding. However, LFSSR-SAV is a CNN-based method, and the limited receptive field of convolutions hinders the utilization of the non-local self-similarity information, especially the inter-view correspondence. 
Therefore, this team introduced the Swin-Transformer to LFSSR-SAV, and designed the spatial-angular versatile Transformer network (namely, SAVformer) for LF image SR. Figure~\ref{fig:vidar} shows the architecture of their SAVformer, which contains Spatial-Former, Angular-Former and EPI-Former. 

\noindent \textbf{Loss Function:} 
To better preserve the geometric consistency, this team followed LF-ATO \cite{ATO} to use the EPI gradient loss $\mathcal{L}_{e}$ and the $\mathcal{L}_{1}$ loss for network training, i.e.,
\begin{equation}
	\mathcal{L}_{total}= \mathcal{L}_{1}+\alpha\mathcal{L}_{e},
	\label{eq:warp}
\end{equation}
where $\alpha$ denotes the weighting factor which is set to 0.1.

\begin{figure*}[t]
     \centering
     \includegraphics[width=0.8\linewidth]{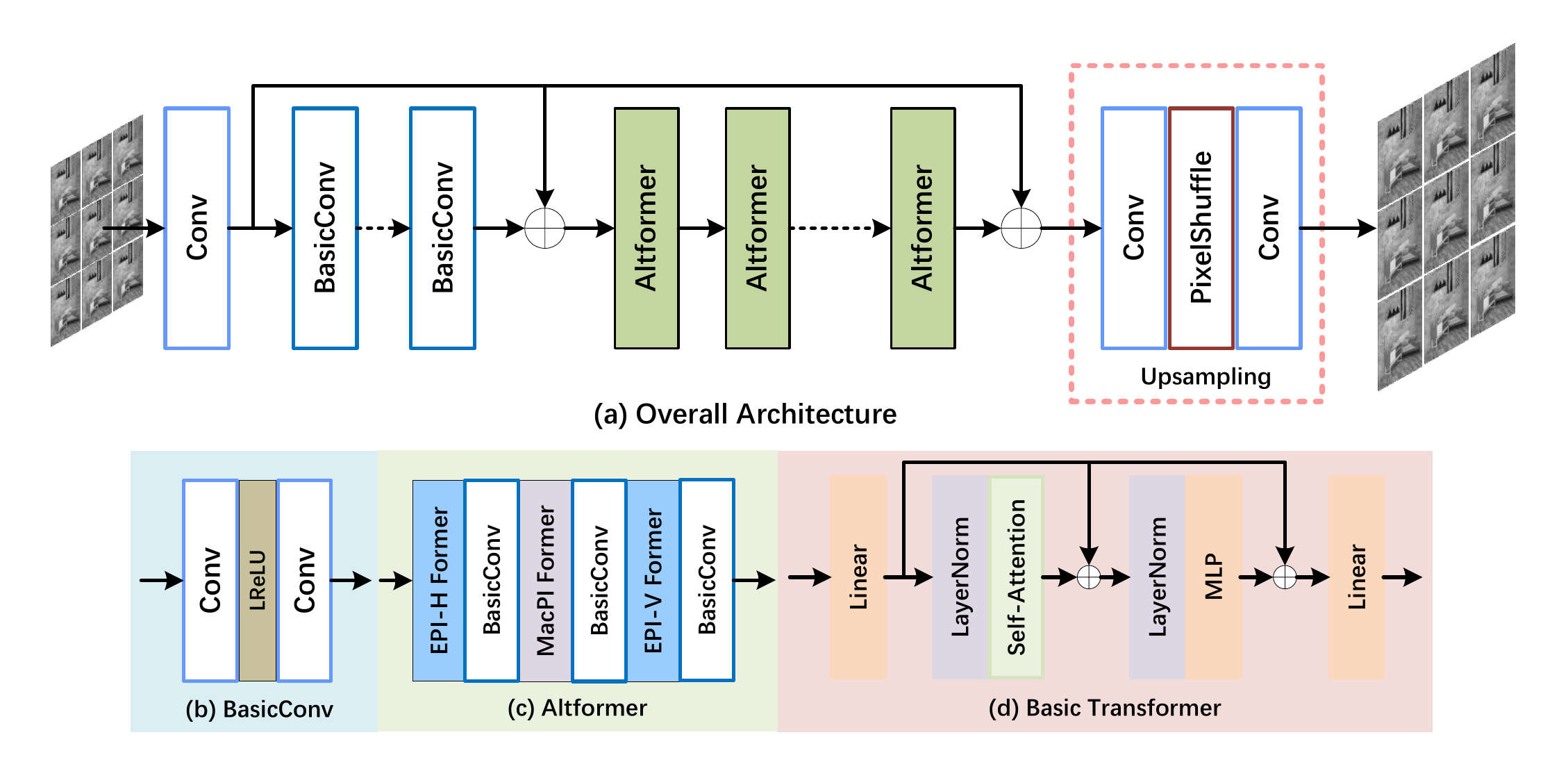}
     \caption{The IIR-Lab Team: The network architecture of the proposed SA-Altfromer.}
     \label{fig:SA-Altfromer}
 \end{figure*}

\noindent \textbf{Training Strategies:}
They trained their network in four stages: 
1) They first trained SAVformer with a batch size of 4, a patch size of $32\times32$, and loss $\mathcal{L}_{total}$ for 16000 epochs (144 iterations per epoch). The learning rate was initially set to $2\times10^{-4}$ and decreased by a factor of 0.5 for every 5000 epochs. 
2) They finetuned SAVformer with a batch size of 4, a patch size of $48\times48$, and loss $\mathcal{L}_{total}$ for 5000 epochs (144 iterations per epoch). The learning rate was initially set to $1\times10^{-4}$ and decreased by a factor of 0.5 for every 1000 epochs.
3) They finetuned SAVformer with a batch size of 8, a patch size of $32\times32$, and loss $\mathcal{L}_{total}$ for 10 epochs (9039 iterations per epoch). The learning rate was initially set to $2\times10^{-5}$ and decreased by a factor of 0.5 for every 15 epochs. 
4) They finetuned SAVformer with a batch size of 4, a patch size of $32\times32$, and loss $\mathcal{L}_{1}$ for 50 epochs (9039 iterations per epoch). The learning rate was initially set to $2\times10^{-5}$ and decreased by a factor of 0.5 for every 15 epochs. 

\subsection{IIR-Lab: SA-Altfromer}

Considering the EPIs and MacPIs are the two typical LF representations that reflect the angular correlations, this team applies Transformers on these two representations to exploit the spatial and angular correlations.
An overview of the proposed SA-Altfromer is shown in Fig.~\ref{fig:SA-Altfromer}(a). 

The proposed method first cascades several BasicConv blocks (as shown in Fig.~\ref{fig:SA-Altfromer}(b)) to gradually extract the intra-view features (i.e., spatial correlations). 
Then, this method adopts 6 Altformer modules (see Fig.~\ref{fig:SA-Altfromer}(c)) to alternately perform multi-head self-attention (MHSA) operations on EPI and MacPI subspace. In each Altformer, the horizontal EPI features, vertical EPI features and MacPI features are sequentially fed into the EPI-H, EPI-V, and MacPI Formers.
As shown in Fig.~\ref{fig:SA-Altfromer}(d), the EPI-H, EPI-V, and MacPI Formers are developed on the Basic Transformer modules. After the Altformers, the enhanced feature by local connection is fed into an upsampling block to generate the final super-resolved results.


\subsection{INSIS: SAMSSR}

\begin{figure*}[t]
     \centering
     \includegraphics[width=0.8\linewidth]{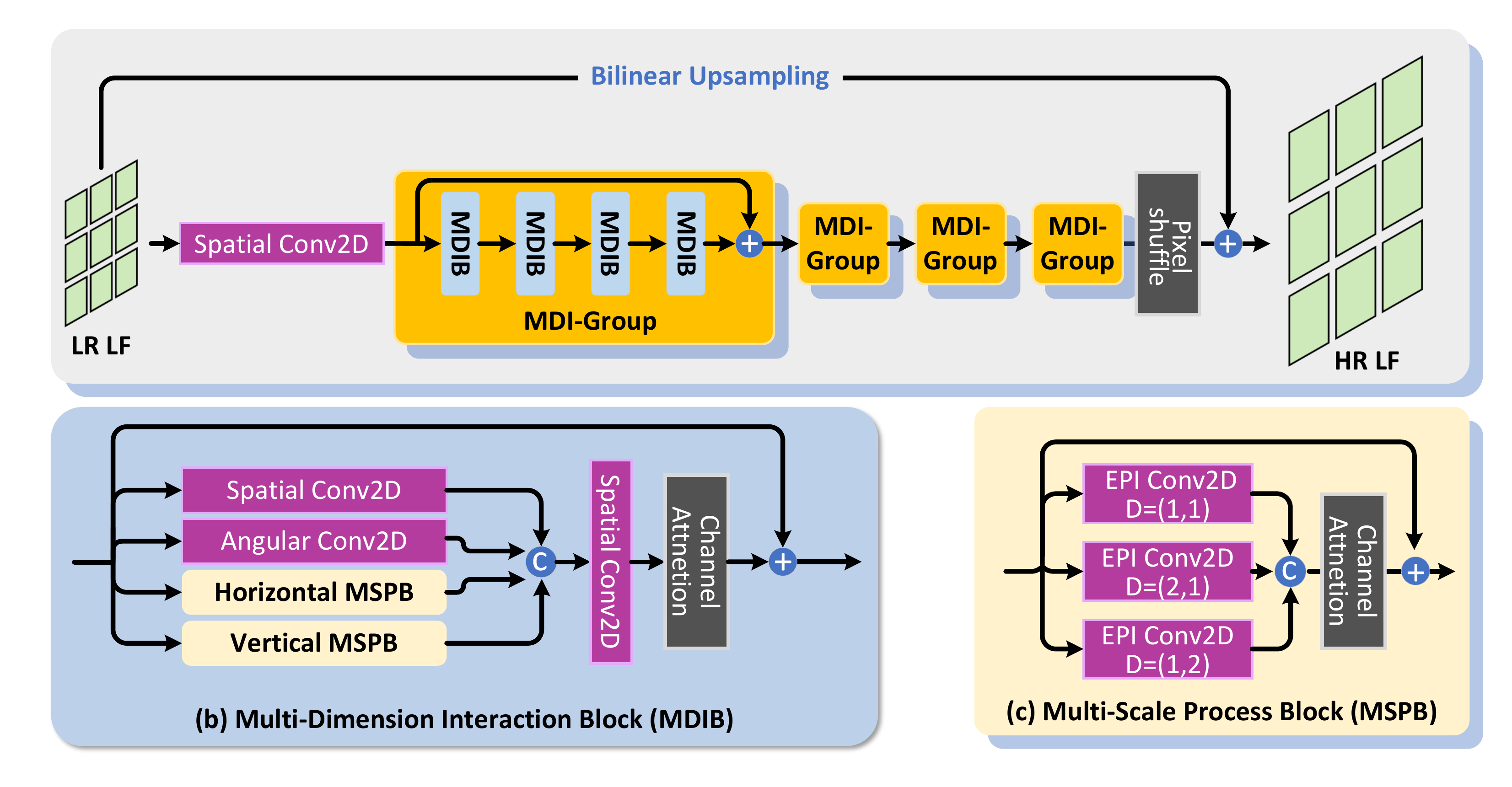}
     \caption{The INSIS Team: The network architecture of the proposed SAMSSR.}
     \label{fig:SAMSSR}
 \end{figure*}

\begin{figure*}[t]
    \centering
    \includegraphics[width=1\linewidth]{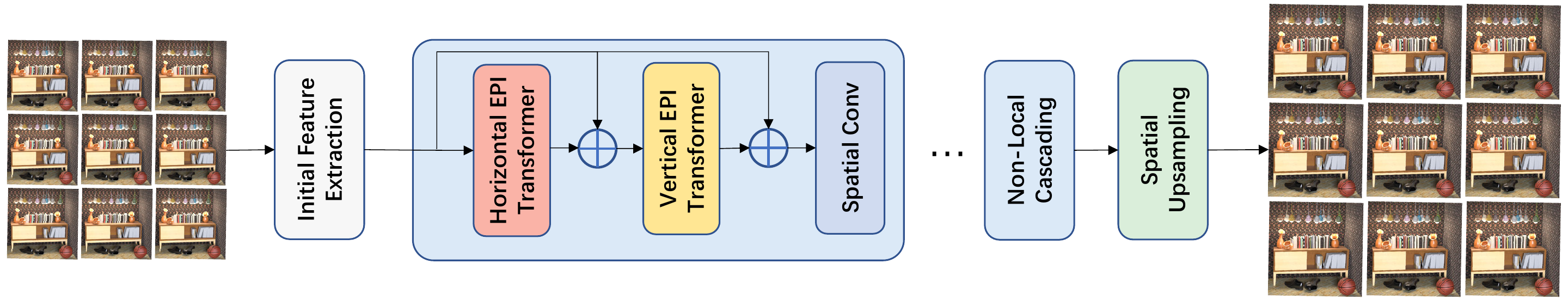}
    \caption{The BNU-AI-TRY Team: The network architecture of the proposed EPITv2\_max.} 
    \label{fig:EPITv2_max}
\end{figure*}

As shown in Fig.~\ref{fig:SAMSSR}, this team proposed a spatial-angular multi-scale spatial SR network (namely, SAMSSR)  to cover the long-range disparity range and explicitly exploit the sub-pixel correspondence in LF images. Readers can refer to \cite{gao2023spatial} for more details of their method.
This team first designed a Multi-Dimension Interaction Block (MDIB) consisting of four branches to separately extract the spatial information, angular information, and horizontal and vertical spatial-angular coupling information. 
To decouple the spatial-angular information along the epipolar line, they designed a Spatial-Angular Multi-Scale Process Module (MSPB) based on horizontal or vertical EPI structures, and adopted dilated convolutions to fully incorporate the long-range disparity information. 
In addition, to better integrate the multi-dimension and multi-scale characteristics, this team adopted the channel attention mechanism at the end of both MDIB and MSPB to fuse information from different branches.

\noindent \textbf{Refinement with Shear Operation.}
To ensure that the proposed SAMSSR performs well under large disparities, this team additionally introduced the LF Shear Attention network \cite{chen2022light} as a second-stage model to improve the accuracy of the final result. 
Specifically, they first applied the pretrained SAMSSR model to the sheared LF images with different disparity values \{-1, -0.5, 0, 0.5, 1\}, and obtained a set of SR results which were then sheared back with the 4$\times$ disparity values \{4, 2, 0, -2, -4\} to restore the original disparity. Afterward, they trained the LF Shear Attention network \cite{chen2022light} to distinguish the relevant information from different sheared levels, and fused them to generate the final SR result.

\subsection{BNU-AI-TRY: EPITv2\_max}

This method is mainly inspired by the recent EPIT method \cite{EPIT}, and aims to improve the capability of the spatial-angular correlation modeling. Specifically, this team increased the channels of feature maps in EPIT (64$\rightarrow$128) and re-designed the non-local cascading block in EPIT by sequentially cascading the horizontal Basic-Transformer unit, the vertical Basic-Transformer unit, and the spatial convolution. 
An overview of their EPITv2\_max is shown in Fig.~\ref{fig:EPITv2_max}.

\noindent \textbf{Data Augmentation.}
During training, this team cropped each SAI into patches of size 128$\times$128 with a smaller stride than EPIT (32 v.s. 64) to generate more LF training patches to alleviate the over-fitting issue. 

\subsection{BIT912: CSWinLFSR}

 \begin{figure*}[t]
     \centering
     \includegraphics[width=1\linewidth]{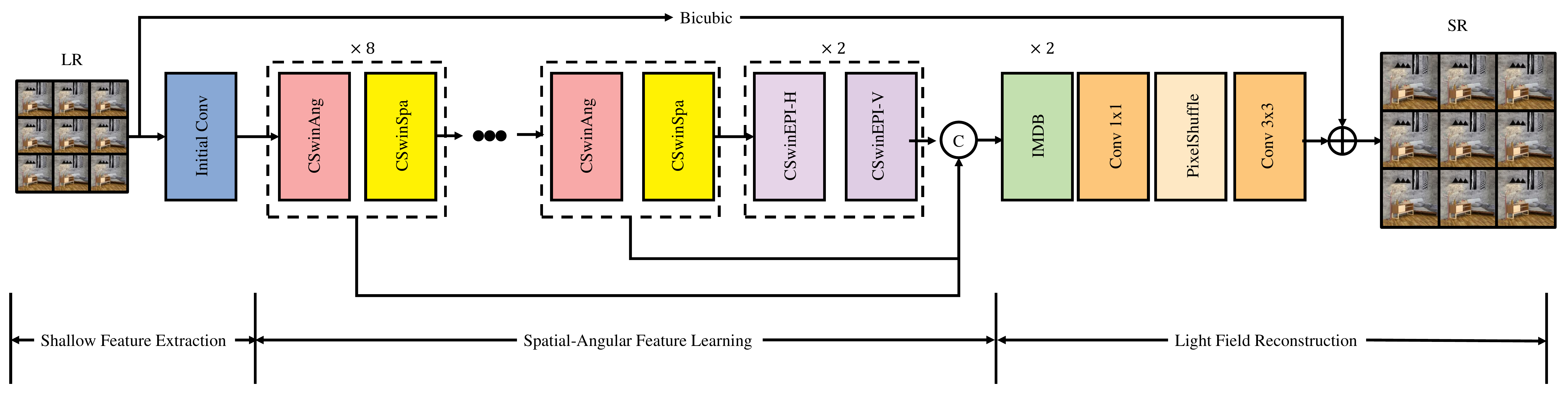}
     \caption{The BIT912 Team: The network architecture of the proposed CSWinLFSR.}
     \label{fig:CSWinLFSR}
 \end{figure*}

This team observed that the Transformer-based method LFT~\cite{LFT} requires a large number of computational resources during LF feature extraction, and thus aimed to reduce the computation cost of LFT and increase the network layers for stronger modeling capability.
Inspired by the novel CSwin Transformer~\cite{dong2022cswin}, this team replaced the global self-attention operation in LFT with the criss-cross shifted window self-attention in CSwin Transformer, and proposed CSwinSpa, CSwinAng, and CSwinEPI modules to extract spatial, angular, and EPI information, respectively. Figure~\ref{fig:CSWinLFSR} shows the overview of the CSwinLFSR network, which consists of three stages: shallow feature extraction, spatial-angular feature learning module, and LF reconstruction. 


\subsection{HawkeyeGroup: LF-DET}

This team proposed a deep efficient Transformers (i.e., LF-DET) for LF image SR. 

\subsection{SHU-IVIPLab: SA-VSNet}

 \begin{figure*}[t]
     \centering
     \includegraphics[width=1\linewidth]{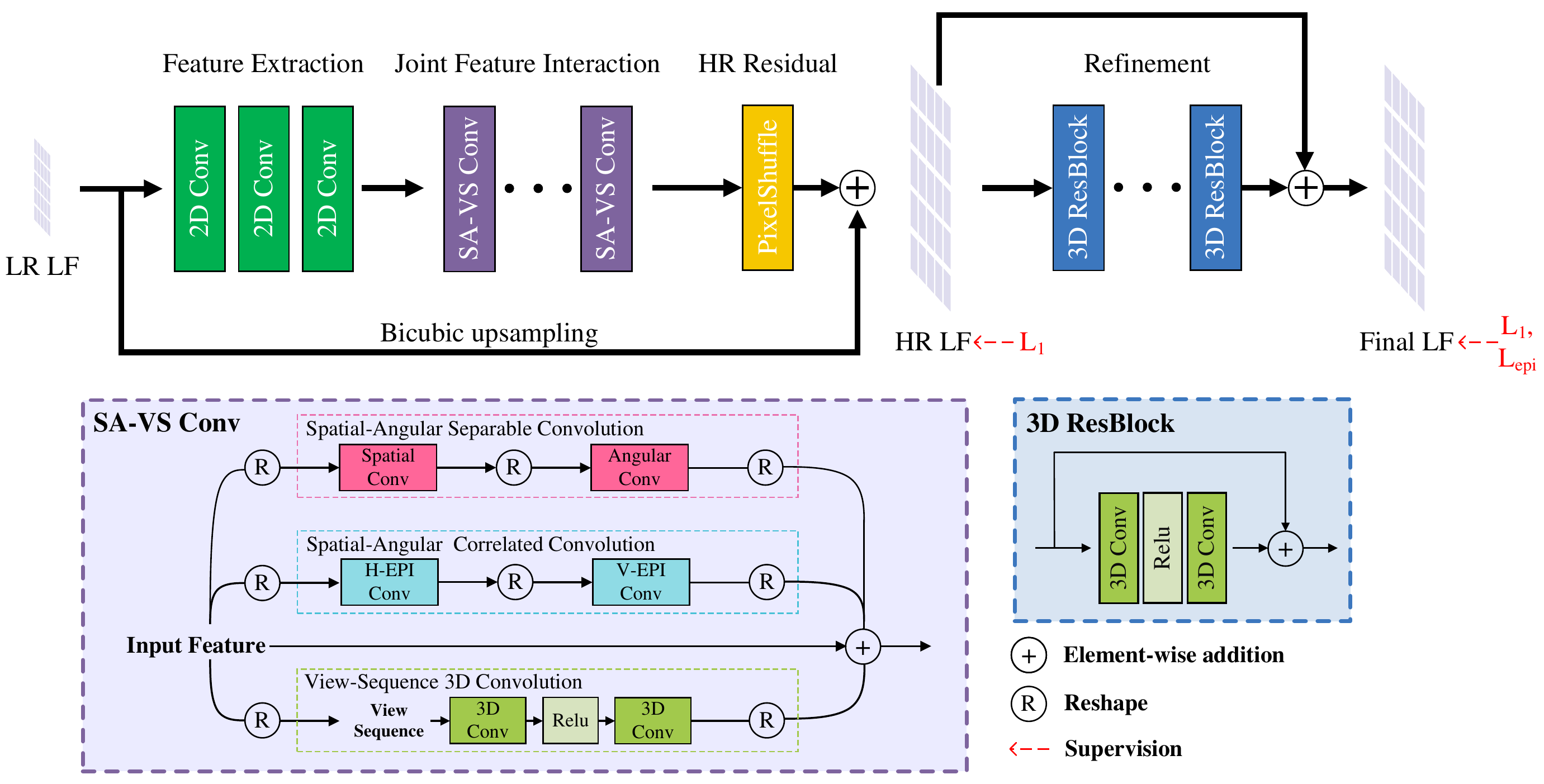}
     \caption{The SHU-IVIPLab Team: The network architecture of the proposed SA-VSNet.}
     \label{fig:SA-VSNet}
 \end{figure*}

Inspired by LFSSR-SAV \cite{LFSSR_SAV}, this team designed a spatial-angular separable convolution (SAS-conv) module and a spatial-angular correlated convolution (SAC-conv) module for LF image processing. This team further introduced 3D convolutions on neighboring view sequences to explore the complementary benefits from the joint spatial context and specific directional views for LF image SR. An overview of proposed Spatial-Angular View-Sequence Network (SA-VSNet) is illustrated in Fig.~\ref{fig:SA-VSNet}.

Specifically, this method follows a ``coarse-to-fine'' strategy to obtain the SR results progressively. 
In the coarse stage, this method first extracts the spatial features from each view of the input LR LF, which are then fed to 16 SA-VS Conv blocks for joint feature interaction. Each SA-VS Conv block consists of an SAS-Conv module, an SAC-Conv module, and a View-Sequence 3D Convolution module. 
The generated features are then processed by a PixelShuffle layer to predict the initial HR LF images which are supervised by the $\mathcal{L}_{1}$ loss.
In the fine stage, this method adopts four 3D residual blocks to further refine the initial super-resolved results, and employs a hybrid loss function consisting of the EPI gradient loss \cite{ATO} and the $\mathcal{L}_{1}$ loss to enhance details.

\subsection{CBNU-MIP-Lab: EPIS-LFSR}

 \begin{figure*}[t]
     \centering
     \includegraphics[width=\linewidth]{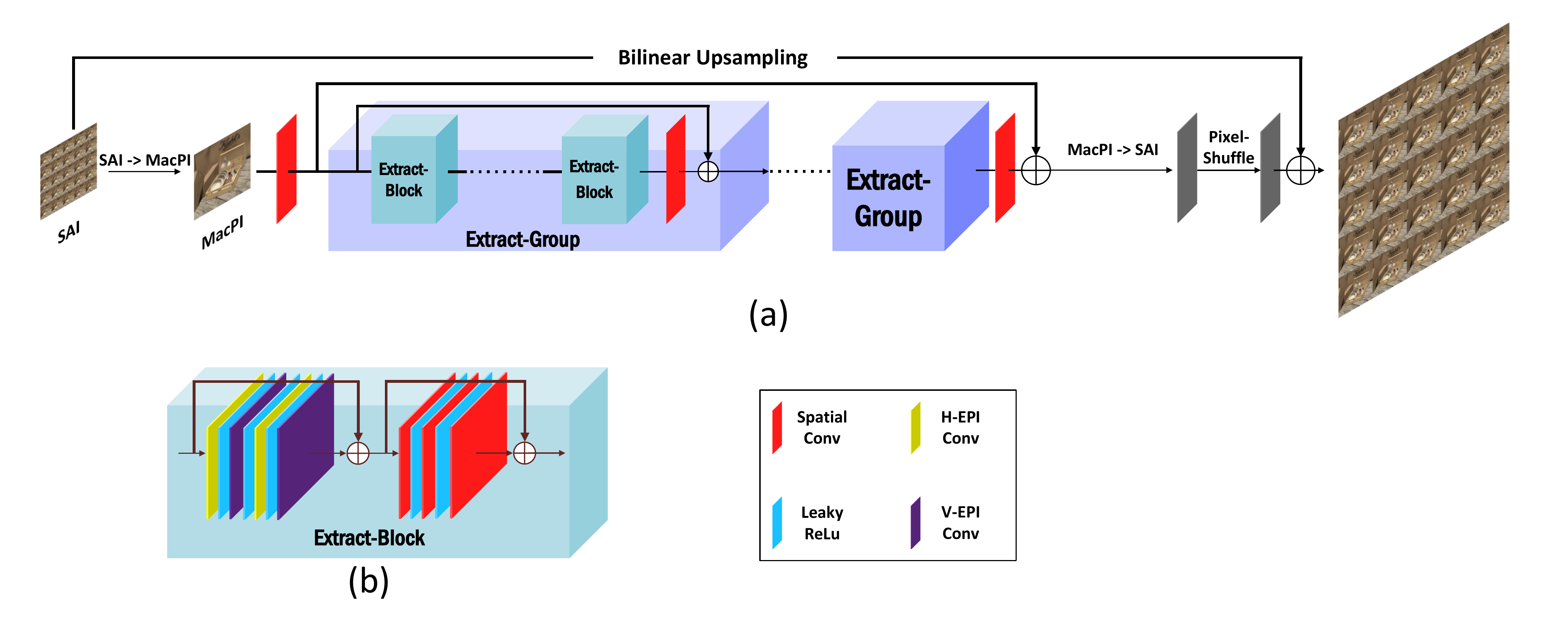}
     \caption{The CBNU-MIP-Lab Team: The network architecture of the proposed EPIS-LFSR.}
     \label{fig:EPIS-LFSR}
 \end{figure*}

Following the pipeline of LF-InterNet \cite{LF-InterNet}, this team rearranged the input LF images into MacPI pattern, and carefully designed a series of 2D convolutions for MacPIs. 
An overview of the proposed network is shown in Fig.~\ref{fig:EPIS-LFSR}, and readers can refer to \cite{salem2023LFSR} for more details of their method.  
The input LR LF is first processed by a spatial convolution to extract shallow features. Then, the shallow features are processed by 8 Extract-Groups (each group consists of 8 cascaded Extract-Blocks) to generate the deep features. 
The proposed network is built in a residual-in-residual manner for better SR performance.

\subsection{LFSR-gdut-team: MAFNetSR}

This team followed DistgSSR \cite{DistgLF} to develop a series of 2D convolutions with channel attention for LF image SR. This method was trained using the default training setting in the BasicLFSR toolbox, and achieved better performance than most baselines.

\section{Acknowledgments}
This work was partially supported by the National Natural Science Foundation of China (No. 61921001, U20A20185, 61972435), the Young Talents Project of Hunan (2020RC3026), the Guangdong Basic and Applied Basic Research Foundation (2022B1515020103), and the Shenzhen Science and Technology Program (No. RCYX20200714114641140, JCYJ20190807152209394). 

    
\section{Teams and Affiliations}
\label{appendix}

\subsection*{Challenge Organizers}
\noindent 
\textbf{\textit{Members:}} \\
Yingqian Wang$^1$ (\href{mailto:wangyingqian16@nudt.edu.cn}{wangyingqian16@nudt.edu.cn}), \\
Longguang Wang$^2$ (\href{mailto:wanglongguang15@nudt.edu.cn}{wanglongguang15@nudt.edu.cn}), \\
Zhengyu Liang$^1$ (\href{mailto:zyliang@nudt.edu.cn}{zyliang@nudt.edu.cn}), \\
Jungang Yang$^1$ (\href{mailto:yangjungang@nudt.edu.cn}{yangjungang@nudt.edu.cn}), \\
Radu Timofte$^{3,4}$ (\href{mailto:timofter@vision.ee.ethz.ch}{timofter@vision.ee.ethz.ch}), \\
Yulan Guo$^{5,1}$ (\href{mailto:guoyulan@sysu.edu.cn}{guoyulan@sysu.edu.cn}).

\vspace{0.2cm}
\noindent 
\textbf{\textit{Affiliations:}} \\
$^1$National University of Defense Technology\\
$^2$Aviation University of Air Force\\
$^3$University of W\"urzburg\\
$^4$ETH Z\"urich\\
$^5$The Shenzhen Campus of Sun Yat-sen University, Sun Yat-sen University\\

\subsection*{(1) The OpenMeow Team}
\noindent 
\textbf{\textit{Members:}} \\
Kai Jin$^1$ (\href{mailto:jinkai@bigo.sg}{jinkai@bigo.sg}), Zeqiang Wei$^{2, 3}$, Angulia Yang$^1$, Sha Guo$^4$, Mingzhi Gao$^1$, Xiuzhuang Zhou$^5$

\vspace{0.2cm}
\noindent 
\textbf{\textit{Affiliations:}} \\
$^1$Bigo Technology Pte. Ltd.  \\
$^2$Smart Medical Innovation Lab, Beijing University of Posts and Telecommunications  \\
$^3$Global Explorer Ltd., Suzhou China \\
$^4$National Engineering Research Center of Visual Technology, School of Computer Science, Peking University  \\
$^5$School of Artificial Intelligence, Beijing University of Posts and Telecommunications

\subsection*{(2) The DMLab Team}
\noindent 
\textbf{\textit{Members:}} \\
Vinh Van Duong$^1$ (\href{mailto:duongvinh@skku.edu}{duongvinh@skku.edu}), Thuc Nguyen Huu$^1$, Jonghoon Yim$^1$, Byeungwoo Jeon$^1$

\vspace{0.2cm}
\noindent 
\textbf{\textit{Affiliations:}} \\
$^1$Department of Electrical and Computer Engineering, Sungkyunkwan University

\subsection*{(3) The VIDAR Team}
\noindent 
\textbf{\textit{Members:}} \\
Yutong Liu$^1$ (\href{mailto:ustclyt@mail.ustc.edu.cn}{ustclyt@mail.ustc.edu.cn}), Zhen Cheng$^1$, Zeyu Xiao$^1$, Ruikang Xu$^1$, Zhiwei Xiong$^1$

\vspace{0.2cm}
\noindent 
\textbf{\textit{Affiliations:}} \\
$^1$University of Science and Technology of China

\subsection*{(4) The IIR-Lab Team}
\noindent 
\textbf{\textit{Members:}} \\
Gaosheng Liu$^1$ (\href{mailto:gaoshengliu@tju.edu.cn}{gaoshengliu@tju.edu.cn}), Manchang Jin$^1$, Huanjing Yue$^1$, Jingyu Yang$^1$

\vspace{0.2cm}
\noindent 
\textbf{\textit{Affiliations:}} \\
$^1$School of Electrical and Information Engineering, Tianjin University

\subsection*{(5) The INSIS Team}
\noindent 
\textbf{\textit{Members:}} \\
Chen Gao$^1$ (\href{mailto:gaochen@bjtu.edu.cn}{gaochen@bjtu.edu.cn}), Shuo Zhang$^1$, Song Chang$^1$, Youfang Lin$^1$

\vspace{0.2cm}
\noindent 
\textbf{\textit{Affiliations:}} \\
$^1$Beijing Key Lab of Traffic Data Analysis and Mining, School of Computer and Information Technology, Beijing Jiaotong University

\subsection*{(6) The BNU-AI-TRY Team}
\noindent 
\textbf{\textit{Members:}} \\
Wentao Chao$^1$ (\href{mailto:chaowentao@mail.bnu.edu.cn}{chaowentao@mail.bnu.edu.cn}), Xuechun Wang$^1$, Guanghui Wang$^2$, Fuqing Duan$^1$

\vspace{0.2cm}
\noindent 
\textbf{\textit{Affiliations:}} \\
$^1$Beijing Normal University \\
$^2$Toronto Metropolitan University

\subsection*{(7) The BIT912 Team}
\noindent 
\textbf{\textit{Members:}} \\
Wang Xia$^1$ (\href{mailto:3220221027@bit.edu.cn}{3220221027@bit.edu.cn}), Yan Wang$^1$, Peiqi Xia$^1$, Shunzhou Wang$^1$, Yao Lu$^{1,2}$

\vspace{0.2cm}
\noindent 
\textbf{\textit{Affiliations:}} \\
$^1$Beijing Institute of Technology \\
$^2$Shenzhen MSU-BIT University

\subsection*{(8) The HAWKEYE Group Team}
\noindent 
\textbf{\textit{Members:}} \\
Ruixuan Cong$^{1,2,3}$ (\href{mailto:congrx@buaa.edu.cn}{congrx@buaa.edu.cn}), Hao Sheng$^{1,2,3}$, Da Yang$^{1,2,3}$, Rongshan Chen$^{1,2,3}$, Sizhe Wang$^{1,2,3}$, Zhenglong Cui$^{1,2,3}$

\vspace{0.2cm}
\noindent 
\textbf{\textit{Affiliations:}} \\
$^1$State Key Laboratory of Virtual Reality Technology and Systems, School of Computer Science and Engineering, Beihang University \\
$^2$Beihang Hangzhou Innovation Institute Yuhang \\
$^3$Faculty of Applied Sciences, Macao Polytechnic University

\subsection*{(9) The SHU-IVIPLab Team}
\noindent 
\textbf{\textit{Members:}} \\
Yilei Chen$^1$ (\href{mailto:yileichen@shu.edu.cn}{yileichen@shu.edu.cn}), Yongjie Lu$^1$, Dongjun Cai$^1$, Ping An$^1$

\vspace{0.2cm}
\noindent 
\textbf{\textit{Affiliations:}} \\
$^1$School of Communication and Information Engineering, Shanghai University

\subsection*{(10) The CBNU-MIP-Lab Team}
\noindent 
\textbf{\textit{Members:}} \\
Ahmed Salem$^1$ (\href{mailto:ahmeddiefy@cbnu.ac.kr}{ahmeddiefy@cbnu.ac.kr}), Hatem Ibrahem$^1$, Bilel Yagoub$^1$, Hyun-Soo Kang$^1$

\vspace{0.2cm}
\noindent 
\textbf{\textit{Affiliations:}} \\
$^1$School of Information and Communication Engineering, Chungbuk National University

\subsection*{(11) The LFSR-gdut-team Team}
\noindent 
\textbf{\textit{Members:}} \\
Zekai Zeng$^1$ (\href{mailto:2112204431@mail2.gdut.edu.cn}{2112204431@mail2.gdut.edu.cn}), Heng Wu$^1$

\vspace{0.2cm}
\noindent 
\textbf{\textit{Affiliations:}} \\
$^1$Guangdong University of Technology

{\small
\bibliographystyle{unsrt}

}

\end{document}